\documentclass[11pt]{article}

\usepackage[preprint]{acl}

\usepackage{times}
\usepackage{latexsym}

\usepackage[T1]{fontenc}
\usepackage{float}

\usepackage[utf8]{inputenc}
\usepackage{booktabs}
\usepackage{multirow}
\usepackage{amsmath}
\usepackage{microtype}

\usepackage{inconsolata}

\usepackage{graphicx}

\title{The Format is the Filter: Scoring Format Fixes What an LLM Judge Can Express}

\author{
  \textbf{Lovedeep Gondara\textsuperscript{*}},
  \textbf{Jekaterina Novikova},
  \textbf{Sagarika Suresh},
  \textbf{Jack Ji},
  \textbf{Jithin Pradeep}
\\
  The Vanguard Group, Inc.\thanks{© 2026 The Vanguard Group, Inc. All rights reserved. \\This material is provided for informational purposes only and is not intended to be investment advice or a recommendation to take any particular investment action.}
\\
  \small{
    \textsuperscript{*}\textbf{Correspondence:}
    \href{mailto:}{Lovedeep\_Gondara@Vanguard.com}
  }
}

\begin{document}
\maketitle
\begin{abstract}
When an LLM judge filters instruction-tuning data, its scoring format
(binary verdict, 1--5 rating, 0--100 score, or comparison) is rarely
justified or ablated. Holding the candidate pool, judge prompt, decoding and
selection budget fixed, we score 15,406 UltraFeedback candidates under five
formats with three judges, curate, fine-tune 66 models, and evaluate them
with four judges. A 0--100 scale is worth only 7--12 equally used levels,
two to three times a 1--5 scale rather than twenty, and a binary verdict
leaves the top score tied in 77--78\% of prompts, so most selections are
made at random. These properties are set by the format almost independently
of the judge. Kendall's $\tau_b$ hides them: it ranks binary best while
$\tau_a$ and top-1 agreement rank it last, because it discounts the ties a
selector must break at random. Both comparative formats inherit a slot bias
whose sign differs across judges. Downstream, curated data beat random
selection under every evaluation judge, robust to length and markdown
adjustment, but no ordering among formats is detectable. We propose
four quantities that judge-based curation pipelines should report.
\end{abstract}

\section{Introduction}
A practitioner assembling an instruction-tuning corpus without an annotation
budget typically asks an LLM judge to score every candidate, keeps the top
fraction, and fine-tunes on what survives
\citep{zheng2023judgingllmasajudgemtbenchchatbot,chen2024alpagasustrainingbetteralpaca,sachdeva2024traindataefficientllms}.
How to ask for the judge's opinion (ACCEPT/REJECT, a 1--5 rating, a 0--100
score, a comparison) is usually decided once, on grounds of convenience, and
rarely ablated with the selector held fixed. Yet a curation pipeline consumes
scores as an \emph{ordering} and breaks ties at random, so tie structure is a
first-class property of a format: a binary verdict leaves every accepted
candidate indistinguishable, and the choice at the frontier is made by chance.

We vary only the format across five designs, score a fixed pool with three
judges, and follow one judge's scores through curation and LoRA fine-tuning
to 66 models. Our main contributions are as follows:
(i)~a fully crossed study of three judges $\times$ five formats,
showing that the format, not the judge, determines how many ties
the scores contain, a result that holds when any one judge is left out and
when each format's number of available levels is taken into account
(\S\ref{sec:variance});
(ii)~evidence that $\tau_b$ inverts the ordering a selection audit needs
(\S\ref{sec:dissociation});
(iii)~evidence that both comparative formats inherit a judge-specific slot
bias, measured three ways (\S\ref{sec:anchored}); and
(iv)~downstream results under four evaluation judges (\S\ref{sec:downstream}).

\paragraph{Related work.}
Elicitation studies measure formats by agreement with human or reference
ratings \citep{li2026gradingscaleimpactllmasajudge,Novikova_2018,zhuang2024yesnoimprovingzeroshot}
or compare pointwise and pairwise protocols for bias
\citep{tripathi2025pairwise}, without following scores into training.
Data-selection work does follow them, but with one elicitation protocol
\citep{chen2024alpagasustrainingbetteralpaca,cui2024ultrafeedbackboostinglanguagemodels,pang2025improvingdataefficiencycurating,wettig2024qurating}.
\citet{deutsch2023ties} establish that ties distort Kendall-style
meta-evaluation; we show the consequence for selection. Configuration beyond
the judge model is a known variance source
\citep{Zhang_2026,donyehiya2026mediocritykeyllmjudge,norman2026reliabilityvaliditysystematiclargescale,soumik2026judgingjudgessystematicevaluation}.
We add the controlled format ablation, carried through to trained models.

\section{Method}
\label{sec:method}
\paragraph{Pool and reference.}
We draw 4,000 UltraFeedback prompts \citep{cui2024ultrafeedbackboostinglanguagemodels},
yielding 15,406 multi-model responses after a 20--6,000 character filter
($\bar{k}=3.85$ per prompt); 300 further prompts are held out. UltraFeedback's
GPT-4 scores are used as the \emph{reference}. They never enter a prompt as numbers,
the selection decision, or the training data, but they choose the anchor in
the anchored format and the evaluation reference, so we report agreement with
the GPT-4 reference and avoid the term accuracy.

\paragraph{Formats.}
The pointwise formats share one preamble (helpfulness, correctness,
completeness, clarity) and differ only in the final instruction; the
comparative formats use a parallel pairwise prompt with five verdicts mapped
to $v\in\{+2,\dots,-2\}$ (Appendix~\ref{app:method}).
\textbf{Binary} (ACCEPT/REJECT $\to\{0,1\}$), \textbf{Likert-5}
($(v-1)/4$) and \textbf{Numeric-100} ($v/100$) are pointwise.
\textbf{Anchored} compares the candidate (always Response A) with a fixed
per-prompt anchor, the upper-middle candidate by GPT-4 score
\citep{donyehiya2026mediocritykeyllmjudge}. \textbf{Tournament} compares every pair of candidates within a prompt
once. Each verdict $v$ gives Response A a win probability
$\sigma(v)=1/(1+e^{-v})$ (0.88 for a strong win, 0.5 for a tie),
and a candidate's score is its average win probability over its matches,
a soft win rate. Decoding is greedy, with a 16-token verdict
budget and reasoning disabled. Selection uses within-format ranks, which
preserve ties, so each format's tie structure passes unchanged into selection.

\paragraph{Judges, curation and training.}
Three curation judges score the identical pool: \texttt{deepseek-v4-flash}
\citep{deepseekai2026deepseekv4highlyefficientmilliontoken},
\texttt{qwen3.5-9b} \citep{yang2025qwen3technicalreport}, and
\texttt{nemotron-3-nano-30b-a3b}; parse rates are $\ge0.998$ in all fifteen
cells. Curation uses DeepSeek scores only. At retention
$r\in\{10,25,50\}\%$ we keep the highest-scored response per prompt (ties
broken at random), then the top $r$ of these globally. \textsc{Random} keeps
one uniform response per prompt and a uniform fraction $r$ of prompts;
\textsc{Length} applies the same policy to response length; \textsc{Full}
uses all candidates. Each subset fine-tunes \texttt{Qwen3-4B}
\citep{yang2025qwen3technicalreport} with LoRA
\citep{hu2021loralowrankadaptationlarge} and three seeds: seven conditions
at three rates plus \textsc{Full}, a total of 66 models.

\paragraph{Evaluation.}
Each model answers the 300 held-out prompts; each response is compared with
the highest-GPT-4-rated completion in both orders, scoring 1 when both orders
agree on a win, 0 on a loss, and 0.5 otherwise. Four evaluation judges
(\texttt{deepseek-v4-pro}, \texttt{gemini-3.5-flash-lite}, \texttt{grok-4.20},
\texttt{Olmo-3-7B}) are analyzed separately. They resolve different
fractions of comparisons into a win or loss, and every unresolved comparison
scores 0.5, so a judge that resolves fewer pulls all win rates toward 0.5;
levels are therefore comparable within a judge but not across judges. Per judge and rate we run a Friedman test
\citep{friedman1937rank}, paired seed-clustered bootstrap contrasts (10,000
resamples), and Holm correction \citep{holmcorrection1979}, and report how
many judges concur.

\begin{table*}[t]
\centering\small
\setlength{\tabcolsep}{3pt}
\resizebox{\textwidth}{!}{%
\begin{tabular}{l rrrrrrr c rrrrrrr c rrrrrrr}
\toprule
& \multicolumn{7}{c}{DeepSeek-V4-Flash} && \multicolumn{7}{c}{Qwen3.5-9B} && \multicolumn{7}{c}{Nemotron-30B-A3B}\\
\cmidrule{2-8}\cmidrule{10-16}\cmidrule{18-24}
Format & Tie & $D_2$ & TT & Spr & $\tau_b$ & $\tau_a$ & T1 && Tie & $D_2$ & TT & Spr & $\tau_b$ & $\tau_a$ & T1 && Tie & $D_2$ & TT & Spr & $\tau_b$ & $\tau_a$ & T1\\
\midrule
Binary      & .501 & 2.0 & .78 & .636 & \textbf{.548} & .222 & .505 && .511 & 2.0 & .77 & .603 & .517 & .210 & .505 && .578 & 1.7 & .78 & .517 & \textbf{.462} & .156 & .481\\
Likert-5    & .240 & 4.2 & .60 & .685 & .527 & .344 & .582 && .304 & 3.3 & .64 & .680 & \textbf{.533} & .337 & .572 && .261 & 3.8 & .66 & .427 & .357 & .224 & .483\\
Numeric-100 & .143 & 7.0 & .45 & \textbf{.722} & .515 & \textbf{.384} & \textbf{.607} && .101 & 9.9 & .40 & \textbf{.708} & .502 & \textbf{.398} & \textbf{.616} && .121 & 8.2 & .47 & \textbf{.532} & .401 & \textbf{.297} & \textbf{.537}\\
Anchored    & .365 & 2.7 & .23 & .412 & .401 & .315 & .526 && .259 & 3.9 & .36 & .397 & .390 & .308 & .526 && .317 & 3.2 & .67 & .207 & .125 & .086 & .396\\
Tournament  & .072 & 13.8 & .05 & .282 & .336 & .304 & .544 && .036 & 27.7 & .10 & .299 & .335 & .292 & .536 && .051 & 19.7 & .04 & .060 & .036 & .035 & .378\\
\bottomrule
\end{tabular}}
\caption{Scoring-stage diagnostics on the identical 15,406-candidate pool.
\emph{Tie}: probability two random candidates receive the same score;
$D_2=1/\text{Tie}$: effective number of equally used levels.
\emph{TT}: fraction of prompts whose top score is tied, so that the selected
response is drawn at random.
\emph{Spr}: global Spearman with the GPT-4 reference; $\tau_b$, $\tau_a$:
mean within-prompt Kendall; \emph{T1}: rate at which the top-scored response
(one seeded tie-break; the exact expectation differs by at most 0.006) is a
reference-top response; 28\% of prompts have a tied reference top, so
uniform selection scores 0.357. Pairs tied in the reference count in no
$\tau$ numerator. Bold: best format per judge. Distinct-value counts and Somers' $D$ are in
Appendix~\ref{app:extra}, Table~\ref{tab:extra}.}
\label{tab:diag}
\end{table*}

\section{Scoring-Stage Results}
\label{sec:judgebehaviour}

\subsection{Format Fixes Score Structure}
\label{sec:variance}
For each diagnostic in Table~\ref{tab:diag} we decompose the sum of squares
over the 15 format--judge cells into format, judge and residual shares
(Table~\ref{tab:anova}); with one observation per cell, the residual absorbs
the interaction.

\begin{table}[t]
\centering\small
\setlength{\tabcolsep}{3.5pt}
\begin{tabular}{lrrrr}
\toprule
& \multicolumn{3}{c}{All three judges} & w/o Nem.\\
\cmidrule(lr){2-4}\cmidrule(lr){5-5}
Diagnostic & Fmt\,\% & Judge\,\% & Res.\,\% & Judge\,\%\\
\midrule
Tie probability            & 97.0 & 0.4  & 2.6  & 0.5 \\
Distinct values            & 99.1 & 0.2  & 0.6  & 0.0 \\
$D_2$/nominal levels$^\dagger$ & 96.0 & 0.3 & 3.8 & 0.0 \\
\midrule
Global Spearman            & 76.4 & 21.9 & 1.7  & 0.1 \\
Within-prompt $\tau_b$     & 55.6 & 36.0 & 8.5  & 0.4 \\
Within-prompt $\tau_a$     & 36.7 & 49.8 & 13.5 & 0.2 \\
Top-1 agreement            & 38.9 & 49.2 & 11.9 & 0.1 \\
\bottomrule
\end{tabular}
\caption{Variance shares over the format$\times$judge grid. The last column
drops Nemotron (DeepSeek and Qwen only); dropping either other judge leaves
the judge share at 0.0--0.5\% (structural) and 23--50\% (reference
agreement). $^\dagger$Pointwise and anchored formats only. Full
leave-one-judge-out results in Appendix~\ref{app:extra}, Table~\ref{tab:lojo}.}
\label{tab:anova}
\end{table}

On structural diagnostics, format explains 97--99\% and the judge under
0.5\%. This is partly by construction: a $K$-level alphabet has tie
probability at least $1/K$, and binary sits almost on its floor of 0.50. The
informative question is how efficiently each judge uses its alphabet.
Normalized by nominal capacity, $D_2$ is still 96\% format (log-normalized,
85\%): every judge uses binary and Likert-5 at 66--100\% of capacity and
numeric-100 at 7--10\% (\S\ref{sec:granularity}). The residual is not
empty: tie probability above the $1/K$ floor is 54\% format and 41\%
residual, so how a judge fills an alphabet is judge-and-format specific.

\paragraph{Reference agreement depends on the weakest judge.}
The judge accounts for 22--50\% of reference agreement in the full grid, but
leave-one-judge-out shows this is carried by Nemotron: with DeepSeek and Qwen
alone, format explains 99.0--99.7\% and the judge 0.1--0.4\%. Given a format, DeepSeek and Qwen give nearly identical agreement, whereas Nemotron lowers agreement under every format. The result does not depend on the two comparative
formats, whose implementations carry slot bias (\S\ref{sec:anchored}):
over the three pointwise formats alone, DeepSeek and Qwen give a format
share of 98.7--99.5\% for $\tau_a$, Somers' $D$ and top-1 (93\% for global
Spearman; Appendix~\ref{app:extra}, Table~\ref{tab:pointwise}). With three judges we do not generalize
this to a capability law.

\subsection{Nominal Granularity is Not Effective Granularity}
\label{sec:granularity}
Numeric-100 offers twenty times the levels of Likert-5, but judges emitted
only 33--37 of its 101 values and placed 74.7\%, 95.9\% and 99.6\% of their
mass (Nemotron, Qwen, DeepSeek) on multiples of five. Counting distinct
values overstates resolution, because a value used twice counts the same as
one used nearly 3,800 times. The effective number of equally used levels,
$D_\alpha=\exp(H_\alpha)$ for R\'enyi entropy $H_\alpha$, corrects this:
numeric-100 is worth 9.1--12.4 levels at $\alpha=1$ and 7.0--9.9 at
$\alpha=2$, against 3.3--4.4 for Likert-5, two to three times the effective
resolution rather than twenty. $D_2$ is exactly the inverse tie probability,
so the tie column of Table~\ref{tab:diag} already reports it. It is also
stable where counts are not: at a third of the pool, distinct values fall
from 37 to $28.0\pm2.0$ while $D_2$ moves by about 0.1 s.d.\ or less
(Appendix~\ref{app:granularity}, Table~\ref{tab:granstab}).

\subsection{Kendall's $\tau_b$ Misleads a Selection Audit}
\label{sec:dissociation}
By $\tau_b$, binary is the best format under DeepSeek and Nemotron and second
under Qwen; by $\tau_a$ it is last under DeepSeek and Qwen. $\tau_a$,
Somers' $D$ and top-1 agreement all rank numeric-100 first, Likert-5 second
and binary last under DeepSeek and Qwen, differing only on anchored versus
tournament. $\tau_b$ \citep{kendalltau1939} divides by
$\sqrt{(n_0-n_1)(n_0-n_2)}$, where $n_1$ counts judge-tied pairs, so coarse
formats are divided by a smaller number. When the reference orders a pair
strictly and the judge ties it, the selector breaks the tie at random, so
the pair's expected concordance is zero, which is how $\tau_a$ scores it and
not how $\tau_b$ does. For example, with reference order $a>b>c$ and a binary
judge accepting $a,b$, $\tau_a=2/3$ and $\tau_b\approx0.82$, yet the selector
picks $a$ only half the time. This is the common case here: binary leaves the
top score tied in 77--78\% of prompts (TT in Table~\ref{tab:diag}), against
40--47\% for numeric-100. The argument extends \citet{deutsch2023ties} from
meta-evaluation to selection.

\subsection{Comparative Formats Inherit Slot Bias}
\label{sec:anchored}
Read each comparative verdict as a margin in favor of Response A. A genuine
preference for the candidate follows the candidate, so its sign flips when
the candidate moves from slot A to slot B; a preference for a slot does not
move. Running 500 anchored candidates per judge in both orders therefore
gives $v_\text{fwd}=q+b$ and $v_\text{rev}=-q+b$, separating the mean
preference for the candidate $q$ from the slot-A bias $b$
(Table~\ref{tab:slot}, Appendix~\ref{app:anchored}). The bias differs in sign on identical inputs:
$b=-0.33$ ($\pm0.09$) for DeepSeek and Qwen, toward slot B, and $+0.89$
[$0.79$, $0.99$] for Nemotron. Self-comparisons confirm it without the additive
model: the anchor is one of the candidates, so 4,000 queries compare a
response with itself, where any verdict other than TIE can only reflect
position. DeepSeek and Qwen return TIE on 95.8\% and 99.6\% of these,
Nemotron on 42.3\%, with a mean verdict of $+0.92$ toward slot A. Because $q$ is a mean preference rather than
discrimination, we also correlate each item's $q_i$ with its GPT-4 gap to the
anchor: $0.44$ [$0.37$, $0.51$] for DeepSeek, $0.37$ [$0.30$, $0.44$] for
Qwen, and $0.10$ [$0.01$, $0.19$] for Nemotron, whose anchored scores are
mostly position.

The tournament inherits the bias through scheduling: pairs were enumerated in
listing order, so the first-listed candidate is always Response A and the
last always Response B. Listing order is unrelated to quality (Spearman 0.05
with GPT-4 score), yet it drives selection in each judge's direction.
DeepSeek selects the last-listed response in 44.6\% of four-candidate
prompts, Nemotron the first-listed in 67.5\%, against 25\% without a position
effect. The tournament's low agreement (global Spearman 0.06--0.30)
is therefore at least partly due to slot bias amplified by unbalanced scheduling.

\section{Downstream Consequences}
\label{sec:downstream}

\begin{table}[t]
\centering\small
\setlength{\tabcolsep}{3pt}
\begin{tabular}{clrrrrr}
\toprule
& $r$ & Contrast & DS-Pro & Gemini & Grok & Olmo \\
\midrule
\multirow{3}{*}{4/4}
 & 50 & anch. $>$ rand.    & $+.093^*$ & $+.086^*$ & $+.084^*$ & $+.049^*$ \\
 & 50 & Lik. $>$ rand.    & $+.099^*$ & $+.087^*$ & $+.110^*$ & $+.054^*$ \\
 & 50 & num. $>$ rand.     & $+.080^*$ & $+.088^*$ & $+.091^*$ & $+.038^*$ \\
\midrule
\multirow{5}{*}{3/4}
 & 25 & Lik. $>$ rand.    & $+.065^*$ & $+.077^*$ & $+.068^*$ & $+.027$ \\
 & 25 & num. $>$ rand.     & $+.061^*$ & $+.094^*$ & $+.072^*$ & $+.009$ \\
 & 50 & bin. $>$ rand.      & $+.048^*$ & $+.086^*$ & $+.082^*$ & $+.026$ \\
 & 50 & Lik. $>$ len.    & $+.062^*$ & $+.079^*$ & $+.078^*$ & $+.013$ \\
 & 50 & tourn. $>$ rand.  & $+.064^*$ & $+.053$   & $+.058^*$ & $+.047^*$ \\
\midrule
\multirow{2}{*}{1/4}
 & 50 & bin. $<$ Lik.    & $-.052^*$ & $-.002$ & $-.028$ & $-.029$ \\
 & 50 & bin. $<$ num.     & $-.032^*$ & $-.003$ & $-.009$ & $-.013$ \\
\bottomrule
\end{tabular}
\caption{Downstream contrasts by the number of evaluation judges concurring
(all 3/4 and 4/4 contrasts; two of the seven 1/4 contrasts). Cells are mean
win-rate differences; $^*$: $p_\text{Holm}<0.05$ for that judge
(seed-clustered bootstrap).}
\label{tab:concur}
\end{table}

The omnibus Friedman test detects differences among the training conditions at
every retention rate under every evaluation judge ($p\le0.017$). Every
contrast that replicates across judges is a curated format beating a
baseline (Table~\ref{tab:concur}): three hold under all four evaluation judges
and five more under three. A difference of $+0.09$ means winning against the
reference on about nine more prompts in 100 than \textsc{Random}. Against \textsc{Full}, trained on all 15,406
candidates, a 400-example curated subset ($r=10\%$) is at least as good:
18 of 20 format$\times$judge differences are positive (up to $+0.059$),
though only 2 survive Holm correction, and \textsc{Random} at 10\% stays
within 0.02 of \textsc{Full}.

\paragraph{Shared judge taste.} Curated students write 1,044--1,257
characters with markdown in 38--46\% of responses, against 1,072--1,113 and
36--40\% for \textsc{Random}. Adding log length relative to the reference and
a markdown indicator to a prompt-fixed-effects regression changes every
curated-versus-\textsc{Random} contrast by less than 0.02
(Appendix~\ref{app:extra}, Table~\ref{tab:lenadj}), so the gains are not a length or markdown effect.

\paragraph{Different data, similar students.} The formats do not select the
same examples. The Jaccard overlap between two formats' DeepSeek-curated
subsets averages 0.02 at $r=10\%$, 0.06 at 25\% and 0.15 at 50\%, so each
format trains on a largely disjoint set. That such different subsets yield
statistically indistinguishable students suggests that, in this pool, many
selections clear the bar that separates curated from random data, and the
large scoring-stage differences are compressed by training.

\paragraph{Between formats.} No format beats another under more than one
evaluation judge. Binary does trail the finer formats in direction, below
Likert-5 in all 12 judge$\times$rate comparisons and below numeric-100 in 11
of 12, matching its weaker scoring-stage agreement. These gaps are too small
for the design to resolve: with 300 prompts and three seeds it detects a
difference of about 0.03--0.05 win-rate points with 80\% power, and its 95\%
CIs rule out format differences larger than 0.064--0.113, depending on the
judge. The missing ordering should therefore be read as undetected; it is not evidence that the formats are equivalent.

\section{Conclusion}
\label{sec:discussion}
Under a fixed decoding rule, the format sets a judge's tie structure almost
independently of the judge, and tie structure decides how often selection is
random; comparative formats also import a judge-specific slot bias, and the format alone largely changes which examples are selected. We
propose reporting the following: \textbf{(1)} the format verbatim, with the decoding rule;
\textbf{(2)} tie probability (or $D_2$), with any distinct-value count given
alongside the pool size; \textbf{(3)} for comparative formats, how positions
were swapped or balanced, and the estimated $b$; and \textbf{(4)} the verdict
parse rate. In practice, we recommend Likert-5 or numeric-100 over a binary verdict,
swap and balance slots in comparative formats, and validate formats with
$\tau_a$, Somers' $D$ or top-1 agreement rather than $\tau_b$.

\section*{Limitations}
\textbf{Decoding.} All verdicts use greedy decoding. Probability-weighted
scores (e.g., the expected Likert value under the verdict-token distribution,
as in G-Eval; \citealp{liu2023gevalnlgevaluationusing}) or repeated sampling
would make binary and Likert-5 effectively continuous. Our structural
results describe format together with greedy decoding, not format alone.
\textbf{Variance decomposition.} There is one observation per cell, so the
residual absorbs the format$\times$judge interaction; three judges give the
judge factor two degrees of freedom, and the judge share in reference
agreement is driven by one judge (Table~\ref{tab:anova}).
\textbf{Reference.} Agreement is measured against UltraFeedback's GPT-4
scores, not human labels. The same scores choose the anchor shown to the
judge in the anchored format and the evaluation reference.
\textbf{Downstream design.} Training data come from one curation judge
(DeepSeek), so we do not test whether a less accurate curation judge yields
worse students; one Nemotron-curated condition would. We have no untuned
student or GPT-4-oracle selector baseline, so we cannot say whether better
reference agreement yields better training data. Our length and markdown
adjustment conditions on properties of the trained model's own outputs, so
it is a sensitivity check rather than a causal control. The \textsc{Length}
baseline is confounded with truncation (the 1,024-token cap removes
11.4--25.2\% of its examples against 1.8--6.2\% elsewhere).
\textbf{Format-specific confounds.} Binary's disadvantage is inseparable from
its random tie-breaking. Anchored is intentionally unswapped, and 26\% of its
queries compare the anchor with itself, which assigns every anchor the same
score. The tournament's slot assignment follows listing order, so its
selections, including the DeepSeek-curated training data, carry each judge's
slot bias; its soft win rate is computed within small slates
($\bar{k}=3.85$), so a weak slate's winner can score as high as a strong
slate's.
\textbf{Scope.} We only use one candidate pool, one student, and one language.

\bibliography{custom}

\appendix
\begin{figure*}[t]
\centering
\includegraphics[width=\linewidth]{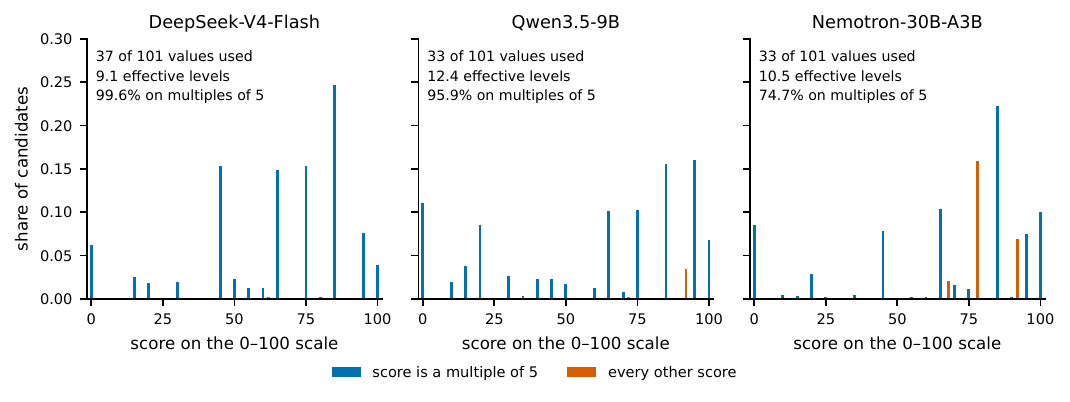}
\caption{Numeric-100 score distributions, one panel per curation judge; blue
bars are multiples of five. Nemotron departs most from the round-number grid
(74.7\% on multiples of five) yet has the lowest reference agreement.}
\label{fig:granhist}
\end{figure*}

\section{Method Details}
\label{app:method}
\paragraph{Prompts.} Pointwise formats:
\begin{quote}\small\ttfamily
You are evaluating the quality of a response to an instruction. Consider
helpfulness, correctness, completeness, and clarity.\\[2pt]
Instruction: \{instruction\}\\[2pt]
Response: \{response\}\\[2pt]
\{format line\}
\end{quote}
\noindent with format lines \textbf{Binary}: ``Is this response of
acceptable quality? Reply with exactly one word: ACCEPT or REJECT.''
\textbf{Likert-5}: ``Rate the response on a 1-5 scale (1=very poor, 2=poor,
3=acceptable, 4=good, 5=excellent). Reply with only the number.''
\textbf{Numeric-100}: ``Rate the response on a 0-100 scale, where 0 is
unusable and 100 is flawless. Reply with only the number.''
Comparative formats use ``You are comparing two responses to the same
instruction. Consider helpfulness, correctness, completeness, and clarity.''
followed by the instruction, Response A, Response B, and ``Which response is
better? Reply with exactly one of: A++ (A much better), A+ (A slightly
better), TIE, B+ (B slightly better), B++ (B much better).'' Verdicts map to
$+2,+1,0,-1,-2$.
 
\paragraph{Anchored.} The anchor is the candidate at index
$\lfloor k/2\rfloor$ after sorting the prompt's candidates by GPT-4 score
(mean anchor score 6.93 against 6.45 for all candidates, so $q<0$ is
expected); the candidate is always Response A, and the score is
$(v+2)/4$. Because the anchor is drawn from the candidates, 4,000 of the
15,406 queries compare the anchor with itself.
 
\paragraph{Tournament.} All $\binom{k}{2}\le6$ within-prompt pairs are
judged once, in the order produced by enumerating the prompt's candidates in
listing order, so candidate $i$ is Response A against every later
candidate. A verdict $v$ gives the first response win probability
$\sigma(v)$ and the second $1-\sigma(v)$; a candidate's score is its mean
win probability over its matches.
 
\paragraph{Fine-tuning.} Each model is trained with LoRA (rank 16,
$\alpha=32$, dropout 0.05) for 2 epochs, using AdamW with learning rate $2\times10^{-4}$,
an effective batch size of 8 and bf16 precision. Training sequences are capped at 1,024 tokens. The cap truncates 4.3\% of
the full pool; across curated subsets it ranges from 1.8\% (numeric-100) to
6.2\% (tournament), but reaches 11.4--25.2\% for \textsc{Length}, which
selects the longest responses by design (see Limitations). All training ran on a single NVIDIA RTX 5090. Curation and evaluation
judges were queried through hosted APIs, except
\texttt{Olmo-3-7B-Instruct}, which we ran locally with vLLM.
 
\paragraph{Statistics.} Every format is evaluated on the same 300 prompts,
a repeated-measures design, so we use the Friedman test per judge and rate,
paired bootstrap contrasts resampling both prompts and seeds, and Holm
correction across format pairs within each rate.
 
\section{Anchored Order-Swap Decomposition}
\label{app:anchored}
Verdicts map to a slot-A-oriented margin: ``A much better'' $\to+2$, ``A
slightly better'' $\to+1$, tie $\to0$, ``B slightly better'' $\to-1$, ``B
much better'' $\to-2$. The candidate's preference travels with the candidate,
so its sign flips when the candidate moves to slot B; a position preference
stays with the slot. Hence $v_\text{fwd}=q+b$, $v_\text{rev}=-q+b$,
$q=(v_\text{fwd}-v_\text{rev})/2$, $b=(v_\text{fwd}+v_\text{rev})/2$. The
additive model treats the five verdicts as an interval scale and can be
distorted near the clipped endpoints. The reverse pass uses a seeded sample
of 500 candidates per judge, excluding self-comparisons; 95\% CIs below are
from 10,000 bootstrap resamples of items.
 
\begin{table}[H]
\centering\small
\setlength{\tabcolsep}{2pt}
\begin{tabular}{lrrrr}
\toprule
Judge & $q$ & $b$ & $\rho(q_i,\Delta)$ & Self-TIE\\
\midrule
DeepSeek & $-0.52$ & $-0.33$ & 0.44 & 95.8\%\\
         & {\footnotesize[$-.65,-.39$]} & {\footnotesize[$-.42,-.24$]} & {\footnotesize[.37,.51]} & \\
Qwen     & $-0.16$ & $-0.33$ & 0.37 & 99.6\%\\
         & {\footnotesize[$-.26,-.07$]} & {\footnotesize[$-.42,-.24$]} & {\footnotesize[.30,.44]} & \\
Nemotron & $-0.07$ & $+0.89$ & 0.10 & 42.3\%\\
         & {\footnotesize[$-.14,+.00$]} & {\footnotesize[$+.79,+.99$]} & {\footnotesize[.01,.19]} & \\
\bottomrule
\end{tabular}
\caption{Order-swap decomposition of anchored verdicts ($n=500$ per judge;
95\% CIs below). $q$: preference for the candidate over the anchor net of
position ($<0$: anchor preferred, as expected). $b$: slot bias ($>0$: favors
Response A). $\rho(q_i,\Delta)$: per-item correlation of $q_i$ with the
GPT-4 score gap, i.e.\ whether preferences track quality. Self-TIE: TIE rate
when the anchor is compared with itself ($n=4{,}000$). DeepSeek and Qwen
share a small slot-B bias but track quality; Nemotron is dominated by a
slot-A bias.}
\label{tab:slot}
\end{table}

DeepSeek and Qwen have indistinguishable slot bias, so the data do not support
a monotone relation between capability and position bias; three judges could
not establish one in any case. Prior reports range from at most 0.04 across
frontier judges \citep{soumik2026judgingjudgessystematicevaluation} to above
0.10 in production judges
\citep{norman2026reliabilityvaliditysystematiclargescale}.

\section{Additional Diagnostics}
\label{app:extra}
Table~\ref{tab:extra} gives distinct-value counts and Somers' $D$; Table~\ref{tab:lojo} the leave-one-judge-out shares; Table~\ref{tab:pointwise} the pointwise-only decomposition; and Table~\ref{tab:lenadj} the length and markdown-adjusted contrasts.
 
\begin{table}[H]
\centering\small
\setlength{\tabcolsep}{4pt}
\begin{tabular}{llrr}
\toprule
Judge & Format & Vals & Somers' $D$\\
\midrule
\multirow{5}{*}{DeepSeek} & Binary & 2 & .261\\
 & Likert-5 & 5 & .406\\
 & Numeric-100 & 37 & .457\\
 & Anchored & 5 & .369\\
 & Tournament & 65 & .364\\
\midrule
\multirow{5}{*}{Qwen} & Binary & 2 & .247\\
 & Likert-5 & 5 & .398\\
 & Numeric-100 & 33 & .472\\
 & Anchored & 5 & .361\\
 & Tournament & 66 & .350\\
\midrule
\multirow{5}{*}{Nemotron} & Binary & 2 & .187\\
 & Likert-5 & 5 & .266\\
 & Numeric-100 & 33 & .355\\
 & Anchored & 5 & .098\\
 & Tournament & 56 & .036\\
\bottomrule
\end{tabular}
\caption{Distinct values emitted and mean within-prompt Somers' $D$.}
\label{tab:extra}
\end{table}
 
\begin{table}[H]
\centering\small
\setlength{\tabcolsep}{3pt}
\begin{tabular}{lrrrr}
\toprule
Judge share (\%) & All & $-$DS & $-$Qwen & $-$Nem.\\
\midrule
Tie probability & 0.4 & 0.4 & 0.0 & 0.5\\
$D_2$/nominal$^\dagger$ & 0.3 & 0.4 & 0.2 & 0.0\\
Distinct values & 0.2 & 0.2 & 0.3 & 0.0\\
Global Spearman & 21.9 & 22.8 & 23.7 & 0.1\\
$\tau_b$ & 36.0 & 32.4 & 34.4 & 0.4\\
$\tau_a$ & 49.8 & 47.0 & 50.4 & 0.2\\
Somers' $D$ & 49.1 & 46.1 & 49.4 & 0.2\\
Top-1 (expected) & 49.0 & 46.2 & 49.4 & 0.3\\
\bottomrule
\end{tabular}
\caption{Leave-one-judge-out judge share of the sum of squares, computed from
unrounded values; top-1 uses the exact expectation under random
tie-breaking. Column headers name the judge removed. $^\dagger$Binary,
Likert-5, numeric-100, anchored.}
\label{tab:lojo}
\end{table}

\begin{table}[H]
\centering\small
\setlength{\tabcolsep}{3pt}
\begin{tabular}{lrrrr}
\toprule
& \multicolumn{2}{c}{All three judges} & \multicolumn{2}{c}{DeepSeek + Qwen}\\
\cmidrule(lr){2-3}\cmidrule(lr){4-5}
Diagnostic & Fmt\,\% & Judge\,\% & Fmt\,\% & Judge\,\%\\
\midrule
Tie probability & 97.5 & 0.4 & 98.0 & 0.1\\
Global Spearman & 9.9 & 80.2 & 93.4 & 4.6\\
$\tau_b$ & 7.7 & 80.8 & 52.9 & 19.8\\
$\tau_a$ & 70.3 & 27.5 & 99.4 & 0.0\\
Somers' $D$ & 71.5 & 26.3 & 99.5 & 0.0\\
Top-1 (expected) & 58.0 & 36.4 & 98.7 & 0.2\\
\bottomrule
\end{tabular}
\caption{Variance shares over the three pointwise formats only (binary,
Likert-5, numeric-100), which have no slot-assignment confound.}
\label{tab:pointwise}
\end{table}
 
\begin{table}[H]
\centering\small
\setlength{\tabcolsep}{3pt}
\begin{tabular}{lrrrr}
\toprule
$r=50\%$ & DS-Pro & Gemini & Grok & Olmo\\
\midrule
Binary      & .048/.047 & .086/.083 & .082/.080 & .026/.025\\
Likert-5    & .099/.094 & .087/.083 & .110/.106 & .054/.052\\
Numeric-100 & .080/.081 & .088/.086 & .091/.090 & .038/.038\\
Anchored    & .093/.080 & .086/.083 & .084/.078 & .049/.044\\
Tournament  & .064/.051 & .053/.049 & .058/.052 & .047/.042\\
\bottomrule
\end{tabular}
\caption{Curated minus \textsc{Random} win rate at $r=50\%$, unadjusted / adjusted for
log response length relative to the reference and a markdown indicator
(OLS with prompt fixed effects over seed-level outcomes). The markdown
indicator is 1 if a response contains a heading, a bullet or numbered list
item, bold text, or a code fence. The largest change is 0.013 (at $r=25\%$
it is smaller); dropping numbered lists, adding tables, using headings and
bold only, or using a log count of markdown elements gives 0.009--0.014.}
\label{tab:lenadj}
\end{table}

\section{Effective Granularity}
\label{app:granularity}
For a score distribution $p$, the R\'enyi entropy of order $\alpha$ is
$H_\alpha=\frac{1}{1-\alpha}\log\sum_i p_i^\alpha$ and $D_\alpha=\exp(H_\alpha)$
is the number of equally used levels with the same entropy. $D_1$ is the
exponentiated Shannon entropy; $D_2=1/\sum_i p_i^2$ is the inverse tie
probability, and $D_2\le D_1$. A distinct-value count weights a score emitted
twice the same as one emitted 3,791 times: DeepSeek's four most frequent
numeric-100 scores (85, 75, 45, 65) carry 70.2\% of the mass. Figure~\ref{fig:granhist} shows the numeric-100 distributions, and Table~\ref{tab:granlevels} lists nominal, emitted and effective levels for every pointwise format.

\begin{table}[H]
\centering\small
\setlength{\tabcolsep}{3.5pt}
\begin{tabular}{llrrrr}
\toprule
Judge & Format & Nom. & Dist. & $D_1$ & $D_2$ \\
\midrule
\multirow{4}{*}{DeepSeek}
 & Binary      &   2 &  2 &  2.00 & 1.99 \\
 & Likert-5    &   5 &  5 &  4.44 & 4.17 \\
 & Anchored    &   5 &  5 &  3.17 & 2.74 \\
 & Numeric-100 & 101 & 37 &  9.07 & 6.98 \\
\midrule
\multirow{4}{*}{Qwen}
 & Binary      &   2 &  2 &  1.98 & 1.96 \\
 & Likert-5    &   5 &  5 &  3.76 & 3.28 \\
 & Anchored    &   5 &  5 &  4.34 & 3.86 \\
 & Numeric-100 & 101 & 33 & 12.39 & 9.94 \\
\midrule
\multirow{4}{*}{Nemotron}
 & Binary      &   2 &  2 &  1.85 & 1.73 \\
 & Likert-5    &   5 &  5 &  4.27 & 3.84 \\
 & Anchored    &   5 &  5 &  3.65 & 3.16 \\
 & Numeric-100 & 101 & 33 & 10.55 & 8.23 \\
\bottomrule
\end{tabular}
\caption{Nominal levels, distinct values emitted, and effective levels.}
\label{tab:granlevels}
\end{table}
 
Binary and Likert-5 use close to their nominal capacity. Anchored uses fewer
effective levels than Likert-5 under DeepSeek and Nemotron, consistent with a
verdict distribution concentrated by the fixed anchor and slot bias; under
Qwen the order reverses ($D_1$ 4.34 vs.\ 3.76).

\begin{table}[H]
\centering\small
\setlength{\tabcolsep}{4pt}
\begin{tabular}{lrrr}
\toprule
& DeepSeek & Qwen & Nemotron\\
\midrule
Distinct & $28.0\pm2.0$ & $27.8\pm1.5$ & $27.8\pm1.4$\\
\quad shift & $-9.0$ & $-5.2$ & $-5.2$\\
$D_1$ & $9.05\pm.09$ & $12.36\pm.12$ & $10.52\pm.11$\\
\quad shift & $-.02$ & $-.03$ & $-.03$\\
$D_2$ & $6.98\pm.08$ & $9.93\pm.11$ & $8.22\pm.10$\\
\quad shift & $-.00$ & $-.01$ & $-.01$\\
\bottomrule
\end{tabular}
\caption{Numeric-100 measures at $n/3$ (mean $\pm$ s.d., 2,000 resamples
without replacement) and their shift from the full-pool value.}
\label{tab:granstab}
\end{table}
 
Subsampling to a third of the pool (Table~\ref{tab:granstab}) removes
16--24\% of the distinct-value count and collapses all three judges to 27.8--28.0, erasing the full-pool
difference (37 vs.\ 33). $D_1$ and $D_2$ shift by at most 0.3\% of their
value and preserve the judge ordering; the $D_2$ shift is about 0.1 s.d. or less\
and the $D_1$ shift within 0.3 s.d.

\end{document}